\documentclass[10pt,journal,compsoc,twocolumn]{IEEEtran}
\ifCLASSOPTIONcompsoc
  \usepackage[nocompress]{cite}
\else
  \usepackage{cite}
\fi

\ifCLASSINFOpdf
   \usepackage[pdftex]{graphicx}
   \DeclareGraphicsExtensions{.pdf,.jpeg,.png}
\else
   \usepackage[dvips]{graphicx}
   \DeclareGraphicsExtensions{.eps}
\fi

\usepackage{algorithm}	
\usepackage{algpseudocode} 

\ifCLASSOPTIONcompsoc
  \usepackage[caption=false,font=footnotesize,labelfont=sf,textfont=sf]{subfig}
\else
  \usepackage[caption=false,font=footnotesize]{subfig}
\fi

\usepackage{url}

\begin{document}
%
\title{ALOJA: A Framework for Benchmarking and Predictive Analytics in Big Data Deployments}
%
%
%
%

\author{Josep Ll. Berral$^1$, Nicolas Poggi$^1$, David Carrera$^1$, Aaron Call$^1$,	Rob Reinauer$^2$, Daron Green$^2$\\
		$^1$Barcelona Supercomputing Center - Universitat Polit\`ecnica de Catalunya\\
		$^2$Microsoft Corporation - Microsoft Research\\
		\{josep.berral,nicolas.poggi,david.carrera,aaron.call\}@bsc.es,\{robrein,dagreen\}@microsoft.com}

\IEEEtitleabstractindextext{%
\begin{abstract}
This article presents the ALOJA project and its analytics tools, which leverages machine learning to interpret Big Data benchmark performance data and tuning. ALOJA is part of a long-term collaboration between BSC and Microsoft to automate the characterization of cost-effectiveness on Big Data deployments, currently focusing on Hadoop. Hadoop presents a complex run-time environment, where costs and performance depend on a large number of configuration choices. The ALOJA project has created an open, vendor-neutral repository, featuring over 40,000 Hadoop job executions and their performance details. The repository is accompanied by a test-bed and tools to deploy and evaluate the cost-effectiveness of different hardware configurations, parameters and Cloud services. Despite early success within ALOJA, a comprehensive study requires automation of modeling procedures to allow an analysis of large and resource-constrained search spaces. The predictive analytics extension, ALOJA-ML, provides an automated system allowing knowledge discovery by modeling environments from observed executions. The resulting models can forecast execution behaviors, predicting execution times for new configurations and hardware choices. That also enables model-based anomaly detection or efficient benchmark guidance by prioritizing executions. In addition, the community can benefit from ALOJA data-sets and framework to improve the design and deployment of Big Data applications.
\end{abstract}

}

\maketitle

\IEEEdisplaynontitleabstractindextext


%
\IEEEpeerreviewmaketitle

\IEEEraisesectionheading{\section{Introduction}\label{sec:introduction}}

\begin{figure}[h!tbp]
\centering
\includegraphics[width=\linewidth]{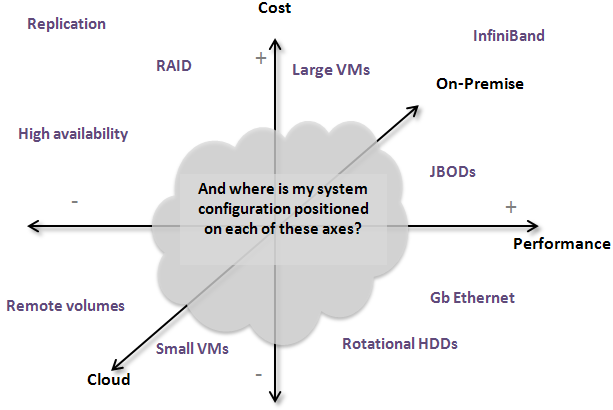}
\caption{Search space for evaluating the cost-effectiveness of a particular set-up}
\label{fig:cloudlocal}
\vspace{-4mm}
\end{figure}

\IEEEPARstart{D}{uring} the last years Hadoop has emerged as the main framework for Big Data processing, and its adoption continues at a compound annual growth rate of 58\%~\cite{person14}. But despite this rapid growth and its expected trend, the Hadoop distributed run-time environment and the large number of deployment choices makes extremely difficult to manage it in a way near to optimal. Hardware component choices and tunable software parameters, from both Hadoop and the Java run-time deployments, have a high impact on performance~\cite{heger:hadoopapproach,Herodotou11starfish:a}. As well as the type of job and the chosen deployment. On-premise deployments or cloud services, produce different behavior patterns during the execution, adding another level of complexity~\cite{DBLP:conf/bigdataconf/PoggiCCMBTAGLRVGB14}.  Figure~\ref{fig:cloudlocal} illustrates the complexity system administrators are faced when planning a new Big Data cluster. Therefore it is usual that Hadoop requires manual, iterative and time consuming benchmarking or tuning, over a huge amount of possible configuration and deployment options. Under this situation any a-priori information provided by an heuristic, oracle or prediction mechanism, providing advice to the decision making process is crucial to improve execution times or reduce running costs.

This article presents the ALOJA framework, its goals towards Hadoop cost-effectiveness analysis, by performing a systematic study of Hadoop deployment variables by benchmarking and tuning Hadoop in different architectures.  Also includes exploration tools for visualization and predictive analytics, with some cases of use for knowledge discovery and Hadoop behavior modeling over benchmarking. ALOJA is an initiative of the Barcelona Supercomputing Center (BSC) in an on-going collaborative engagement with the Microsoft Product groups and Microsoft Research (MSR) to explore upcoming hardware architectures and building automated mechanism for deploying cost-effective Hadoop clusters. The initial approach for ALOJA project was to create a comprehensive open public and vendor-neutral Hadoop benchmarking repository. This benchmark repository is intended to compare software configuration parameters, state of the art and emerging hardware like solid-state disks or RDMA oriented networks such InfiniBand, also different types of Cloud services. Further, the ALOJA project studies the cost-effectiveness for possible set-ups along with their run-time performance, oriented to automate recommendations for configurations given specific workloads, serving also as a reference guide for selecting new infrastructures and designs of Hadoop clusters. 

Once focused on cost-effectiveness studies, the latest phase of the ALOJA project focuses on providing tools to automate both the knowledge discovery process and performance predictive analytics of Hadoop benchmark data. As previously exposed, assisting decision making processes may require manual and time-consuming benchmarking followed by operational fine-tuning for which few organizations have either the time or performance profiling expertise. Finding generalizable rules of thumb for configuring such complex systems, that also can be applied to all workloads (in this case Hadoop jobs), becomes extremely difficult and it is an inherent problem on such type of complex systems.
For this reason automatized modeling processes are of great help finding, if not general, several specific models for each environment or type of workload. Modeling and prediction ALOJA tools provide machine learning methods to create models from past Hadoop executions and use them to predict the behavior of new and unseen executions by their software or hardware configurations and given workload, also used to recommend configurations towards optimizing the performance of such workload. Those tools can also be used to detect anomalous Hadoop execution behaviors, by comparing observed executions with their expected behaviors, using model-based outlier detection algorithms. Further, the created models can be used to describe how each variable influences in the job execution, and prioritize them when choosing a software configuration or a hardware infrastructure.

\subsection{Motivation}

During the past years, most of the industry efforts have focused into building scalable data processing frameworks like Hadoop and its derived services, but those efforts have derived into the adoption of Hadoop by companies, the development of Map/reduce applications, and lately into tuning the performance of such deployments and the management of their data. Studies show that Hadoop execution performance can improve up to three times at least from the default configuration for most deployments~\cite{heger:hadooptuning}.

Hadoop is currently dominated by several vendors, each one offering their customized distribution with patches and changes on the default Apache distribution, thanks to the fact of Hadoop itself being open-source. Such changes are rarely pushed into the main distribution, maintaining its condition of \textit{standard} and default, so running without speed-ups or performance tweaks~\cite{heger:hadooptuning}. There is also evidence that Hadoop performance is quite poor on new or scale-up hardware~\cite{Appuswamy2013}, also scaling out in number of servers tends to improve performance but paying extra running costs like power, also storage space~\cite{Appuswamy2013}. These situations make a case to reconsider how to scale hardware and services from both research and industry perspectives.

Optimizing Hadoop environments requires to have run multiple executions and examine large amount of data to understand how each component affects the system. This data is usually obtained from the environment and infrastructure description, the Hadoop configuration parameters, the outputs from the execution and the performance logs; it can be some times Gigabytes per execution, and examining it manually can be just impossible. This is a challenge for automatic modeling methods like machine learning. 

Discovering which features are the most relevant and which ones provide useless information is key to discover which ones have the biggest impact on performance or which ones can change without affecting the execution, allowing users or operators to focus their attention onto the relevant parameters or infrastructure elements, also adjusting the free features to their available resources. Machine learning techniques provide not only the ability to model a system and predict its behavior given a set-up, but also to explore the environment features and discover knowledge about our system. And from here on, use this knowledge and models as rules, heuristics or oracles to make decisions or make recommendations based on diverse potential goals like increase performance, reduce execution costs.

Our goal is to empower users and operators with the ability to predict workload performance and provide them a clear understanding of the effect of configuration or infrastructure choices.

\subsection{Contribution}

The ALOJA framework for Big Data benchmarking focuses on providing researchers and developers a set of tools and visualizations to characterize Hadoop configurations and performance issues. The framework is available to all the community to be used with own Hadoop data-set executions, also users and researchers can implement or expand this tool-set through adding, comparing or predicting data from observed task executions and/or by adding new analysis or visualization tools, even new machine learning or anomaly detection algorithms. Also, all the data-sets collected for ALOJA are public and can be explored through our framework or used as data-sets in other platforms.

Finally, in this work we share our experiences in analyzing Hadoop execution data-sets. We present results on early findings using our visualization tools. Also we present results on modeling and prediction of execution times for a set of Hadoop deployments and the HiBench benchmarking suite~\cite{HiBench}. Also we present some cases of use for the predictive analytics tools, like our anomalous execution detection mechanisms, a model-based methods for recommending configurations for benchmarking new infrastructures, and a simplistic algorithm for ranking features.

This article is structured as follows: Section~\ref{sec:soa} presents the preliminaries for this work and current state-of-art. Section~\ref{sec:aloja} present the ALOJA framework, the project road-map and its early findings. Section~\ref{sec:machinelearning} presents the machine learning methodology, the data-sets used for predictive analytics, and modeling and prediction results. Section~\ref{sec:casesofuse} presents some cases of use for the prediction models, focused on anomaly detection on Hadoop executions, recommendation of benchmarking new infrastructures, and detecting relevant features when making decisions. Finally, Section~\ref{sec:conclusions} summarizes the paper and provides the conclusions and future work lines for the project.

\section{State of the Art}
\label{sec:soa}

Here we present the background of the ALOJA project, and the current state of the art in Hadoop and distributed systems performance oriented prediction analytics.

\subsection{Background}
\label{sec:background}

The work presented here focuses on the ALOJA project, an initiative of the Barcelona Supercomputing Center (BSC) that has developed Hadoop-related computing expertise for more than 7 years~\cite{bsc:autonomic}. This project is partially supported by the Microsoft Corporation, contributing technically through product teams, financially, and by providing resources and infrastructure as part of its \emph{Azure4Research} program. The initial approach was to create a comprehensive vendor-neutral open and public Hadoop benchmarking repository, currently featuring more than 40.000 Hadoop benchmark executions, used as the main data-set for this work. At this time, the tool compares software configuration parameters, emerging hardware and Cloud services, also costs for each type of presented set-up along with the resulting performance for each workload. We expect our growing repository and analytic tools will benefit Hadoop community to meet their Big Data application needs.

\subsubsection{Benchmarking}

Due to the large number of possible Hadoop configurations, each one affecting the execution in a different way, we are characterizing Hadoop through extensive benchmarking. Hadoop distributions include jobs that can be used to benchmark performance, usually referred as micro benchmarks, each one representing specific types of workloads. However, ALOJA currently features executions from the Intel HiBench open-source benchmark suite~\cite{HiBench}, which can be more realistic and comprehensive than the supplied examples in Hadoop distributions.

\subsubsection{Current Platform and Tools}

The current ALOJA platform is an open-source software and data-sets available for the community to download, use and expand. The goal is to achieve automated benchmarking of Big Data deployments either on-premise or in the Cloud. The framework counts with a set of scripts to automate cluster and node definitions, taking care of describing cluster orchestration and set-ups. Also it contains scripts to execute selected benchmarks and gather results and related information, importing them to the main application featured by a specialized repository with a web interface containing the visualization and predictive analytics tool-sets. Through this web-based repository, the user can explore the imported executions and their details.

\begin{figure}[h!tb]
\centering
\includegraphics[width=\linewidth]{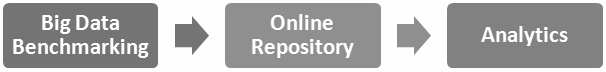}
\caption{Workflow of the ALOJA framework}
\label{fig:workflowaloja}
\end{figure}

Figure~\ref{fig:workflowaloja} shows how the main components of the ALOJA platform feed each other in a continuous loop: benchmarks are executed and the information is collected into the on-line repository, so users can explore and then decide which new executions or benchmarks will follow. One of the goals of the ALOJA project is to automate this process by using data-mining capabilities through machine learning methods, enabling automated knowledge discovery and characterization, and thus recommending which executions are interesting to push into the workflow. The platform includes a Vagrant virtual machine~\cite{vagrant} with a sand-box environment and sample executions used for development and early experiments, for users to create their own repositories or data analysis. In the project site~\cite{bsc:hadoop} there is more technical documentation for further usage and tool development, also preliminary versions of this work and project were presented in~\cite{berral-kdd2015} and \cite{DBLP:conf/bigdataconf/PoggiCCMBTAGLRVGB14}.

\subsection{Related Work}

As previously said, for most deployments, execution performance can be improved by at least 3 times from the default Hadoop configuration~\cite{heger:hadooptuning}, and the emergence of Hadoop in the industry has led to several attempts at tuning towards performance optimization, new schemes for proper data distribution or partition, and adjustments in hardware configurations to increase scalability or reduce running costs. Characterizing these deployments is a crucial challenge towards looking for optimal configuration choices. An option to speed-up computing systems would be to scale-up or add new (and thus improved) hardware, but unfortunately there is evidence that Hadoop performs poorly in such situations, also scaling out in number of servers improve performance but at the increased costs of infrastructure, power and required storage~\cite{heger:hadooptuning}.

Previous research works, like the Starfish Project by H.Herodotou et al.~\cite{Herodotou11starfish:a}, focus on the need for tuning Hadoop configurations to match specific workload requirements. Their work proposed to observe Hadoop execution behaviors, obtaining profiles and using them to recommend configurations for similar workloads. This approach has been a useful reference for ALOJA focusing on modeling Hadoop behaviors from observed executions, but instead of just collecting and comparing behavior features, we apply machine learning methods to characterize those behaviors across a large corpus of profiling data in our predictive analytic tools.

The idea of using machine learning with self-configuring purposes has been seen previously in the field of autonomic computing. Works like J.Wildstrom~\cite{Wildstrom:2007:MLO:1625275.1625456} proposed modeling system behaviors vs. hardware or software configurations, focusing on hardware reconfiguration on large data-center systems. Also other frameworks like the NIMO framework (P.Shivam~\cite{Shivam:2006:AAL:1182635.11641734}) modeled computational-science applications allowing prediction of their execution time in grid infrastructures. These efforts are precedents of successful applications of predictive analytics through machine learning in distributed systems workload management. In the ALOJA framework we are applying such methodologies to complement the exploration tools, allowing the users, engineers and operators to learn about their workloads in a distributed Hadoop environment.

\section{The ALOJA Framework}
\label{sec:aloja}

The project ALOJA is born in the attempt to provide solutions to an important problem for the Hadoop community, that is the lack of understanding of which parameters, either software or hardware, determine the performance and costs of Hadoop workloads. Further, Hadoop deployments can be found in a diversity of operational environments, from low-end commodity clusters to high-end data appliances, and including all types of Cloud-based solutions at scale, resulting in a growing need to understand how different operational parameters, such as VM sizes used in Cloud deployments, affect the cost-effectiveness of Hadoop workloads.

The project is structured in three phases, aiming to 1) create the benchmarking ALOJA platform, 2) deploy visualization and analytic tools in an on-line Web application and 3) develop performance models deriving from the collected data. The information obtained from the developed models and benchmarking information should guide users to choose what deployment options are the most adequate to balance the cost-effectiveness of their workloads.

\subsection{Methodology and Road-Map}

The road-map of the ALOJA project is structured in 3 phases of execution:

\textbf{First phase}: The initial phase consists on perform a systematic study of performance results across a range of selected state of the art and novel hardware components, software parameters configuring Hadoop executions, and solution deployment patterns.

\textbf{Second phase}: The second phase introduces models and methods for analytics of Hadoop executions. The ALOJA repository is an accumulated data-set of results, allowing us to model and predict performance features given an input set of workload execution characteristics, hardware components, software configurations and the deployment characteristics. Predicting these performance and efficiency outcomes given the execution characteristics will also let the system to make decisions on the present and future executions, like discarding executions tagged as anomalous, or recommend the next batch of executions to be performed.


Visualization tools are included into the platform, to observe the large amount of execution results in a more comfortable and comprehensive way, allowing a better understanding of the executions. One of the checkpoints for this phase is to be able to answer questions like \emph{which software and hardware configuration is the best for my Hadoop workload}. All of this by having into account the budget or user indicated hardware limitations.

\textbf{Third Phase}: The third phase of ALOJA focuses on applying the predictive analytic models to automate decisions like deciding which executions submit for current or new Hadoop deployments, having into account cost-effectiveness and user constraints on budget or hardware availability, and validate or reject executions in case of anomalous results or behavior.

\subsection{The platform}

The ALOJA platform implements the previously explained repository and tools, available at the ALOJA web-site~\cite{bsc:hadoop}. The relevant components of this implementation are the benchmark managing, the data analysis and flow, the deep instrumentation, and the testing infrastructure.

\subsubsection{Benchmarking Components}
\label{sec:benchmarks}

The \emph{configuration management} scripts are in charge of setting up the servers (on premise or Cloud), the OS and JVM configuration and profiling tools, the Hadoop deployment plus benchmark launching and metrics collection, also the environment cleanup. The \emph{parameter selection and queuing} builds the batch of workloads to run from the user options introduced through the ALOJA web-application, also schedules them into queues. The \emph{metrics profiling} captures the system metrics for each process and execution (CPU, memory, IO, storage), including also a log parser to match the collected Hadoop metrics with system metrics. Finally the \emph{benchmark suite} includes the set of benchmarks to choose from, featuring the previously cited Intel HiBench Suite characterizing 4 categories of workload: $\mu$-benchmarks, web search, machine learning and HDFS. The benchmarks currently examined in ALOJA are the following:
\begin{itemize}
\item \emph{Terasort}, sorts 1TB of data generated by the TeraGen application (I/O and CPU intensive).
\item \emph{Wordcount}, counts number of word occurrences in a large text files (CPU bound).
\item \emph{Sort}, sorts the input directory into an output directory (I/O intensive).
\item \emph{Pagerank}, crawls Wikipedia sample pages, using the Google’s Web page ranking algorithm.
\item \emph{Bayes}, Bayesian classification using the Mahout library, using a subset of the Wikipedia dump.
\item \emph{K-means}, Mahout’s implementation of the k-means algorithm.
\item \emph{Enhanced DFSIO}, an I/O intensive benchmark to measure throughput in HDFS using map reduce.
\end{itemize}

\subsubsection{Data Analysis and Flow}

After the execution of a benchmark, the result is sent to the repository. Performance metrics are extracted, parsed and joined with the Hadoop metrics for each job, and finally imported into a relational database, where each benchmark is assigned a unique ID and related to its configuration (SW and HW), its system metrics and Hadoop results. Once in the database, the platform provides filtering and visualization for the stored data, allowing to compare different runs, parameter choices, or deployment options.

\subsubsection{Deep instrumentation}

ALOJA provides a mechanism to compare directly results from different Hadoop executions in an easy way, but in some cases it will be necessary to conduct analysis in a lower level. For such analysis ALOJA uses low-profiling tools:
\begin{itemize}
\item \emph{PARAVER}~\cite{bsc:tools}, a performance visualization tool for execution traces developed at the BSC, and widely used for analyzing HPC applications, but also web applications and now for Hadoop workloads
\item The \emph{Hadoop Analysis Toolkit}, including the Java Instrumentation Suite (JIS) that uses Aspect Oriented Programming to avoid recompiling Hadoop, and producing an execution trace file.
\item Advanced analysis tools like DBSCAN~\cite{DBLP:journals/datamine/SanderEKX98} to detect different algorithm phases as well as regions of different subroutines with similar behavior inside Hadoop jobs. This method is going to be hooked to the data flow to enhance the collected system metrics.
\end{itemize}

\subsubsection{Initial Testing Infrastructure}

As ALOJA aims to create a vendor-neutral repository, the execution environment for tests is not a closed specification, and the platform is designed to support multiple deployment environments. However, in order to test the platform, a few initial deployments have been added and passed through benchmarks:

1) A \textbf{High-End Cluster}: an on-premise cluster using 2 to 16 data-nodes per execution plus a head-node, with two 6-core sandy-bridge Intel processors and 64GB of RAM machines; 6 SATA2 SSD drives as RAID-0 storage (1.6GB/s read, 1GB/s write), and local 3TB SATA HDD storage per node; network interfaces with 4 Gigabit Ethernet and 2 FRD InfiniBand ports per node providing a bandwidth peak up to 56Gbps, and InfiniBand switching.

2) A Cloud IaaS: a \textbf{Microsoft Azure} environment, using A7 instances using 3 to 48 data-nodes per execution plus a head-node, with 8 cores and 56GB of RAM; mounting up to 16 remote volumes limited to 500 IOPS. The initial ALOJA repository includes executions with different storage configurations, varying the amount of remote volumes and exchanging remote and local volumes.

\section{Modeling Benchmarks}
\label{sec:machinelearning}

In order to enhance the knowledge discovery capabilities of ALOJA filters and visualization tools, we introduce the predictive analytics component ALOJA-ML. As part of the second and third phase, the ALOJA project aims to include data-mining techniques in the analysis of Hadoop performance data. Modeling the Hadoop behavior towards executions allows to predict such executions output values (e.g., execution time or resource consumption) based on input information like software and hardware configurations. Also, such models can be applied in anomaly detection methods, by comparing actual executions against predicted outputs, also tagging as anomalous those tasks whose run-time lies notably outside their machine-learned prediction. Furthermore, through data-mining techniques we can find which data from our repository has more significance, identifying which minimal set of executions is required to characterize a specific Hadoop deployment, and then being able to recommend which executions should be performed to benchmark a new or updated deployment.

\begin{figure}[h!tb]
\centering
\includegraphics[width=\linewidth]{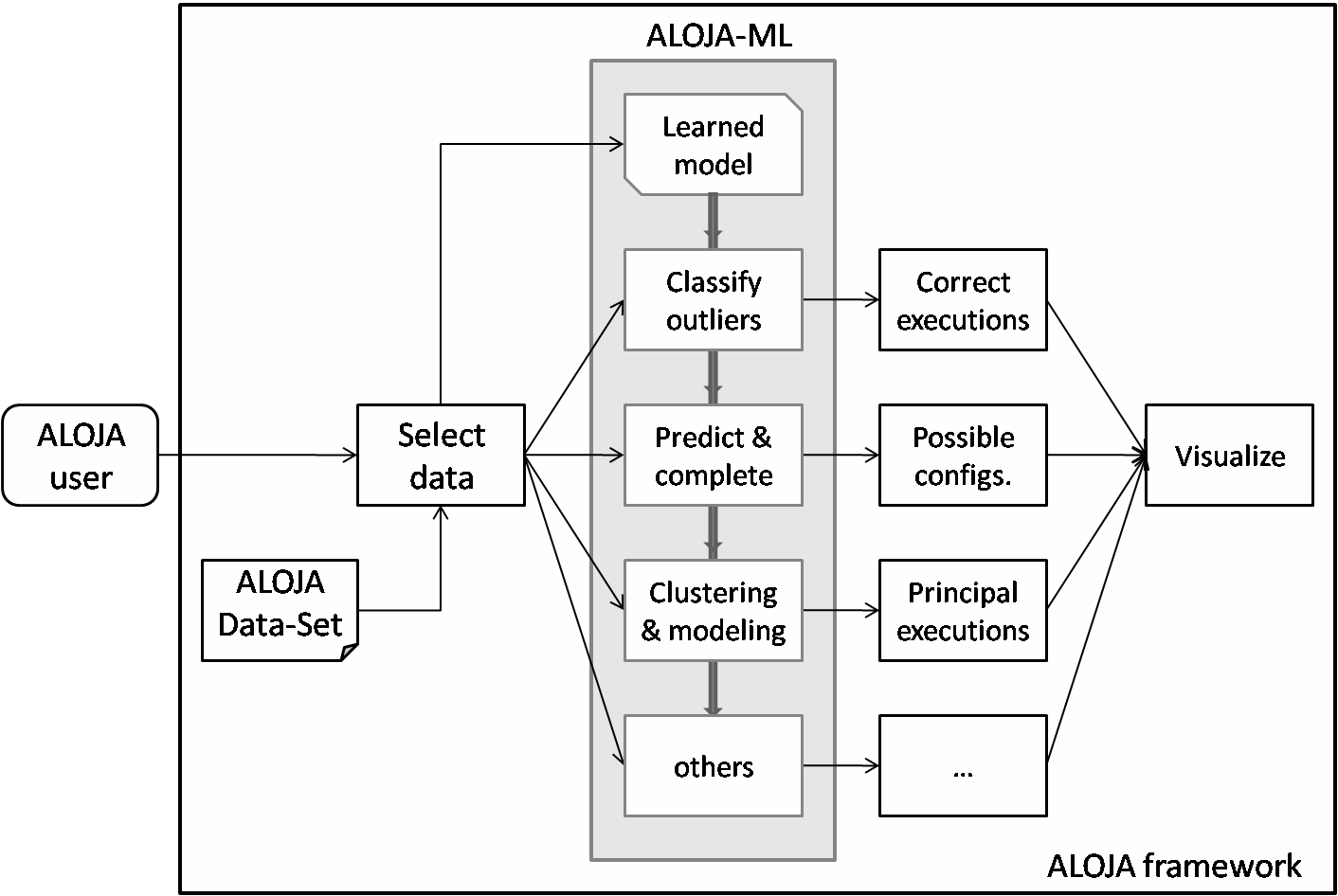}
\caption{ALOJA-ML, the predictive analytics add-on component of ALOJA}
\vspace{-2mm}
\label{fig:component_alojaml}
\end{figure}

While the ALOJA framework collates and analyzes data collected from Hadoop task executions and displays it through a range of tools, helping users understand and interpret the executed tasks, ALOJA-ML complements this by adding tools that learn from the data and extract hidden (or not so obvious) information, also adding an intermediate layer of data treatment to complement the other visualization tools. Figure~\ref{fig:component_alojaml} shows the role of ALOJA-ML inside the ALOJA framework.

\subsection{Data-Sets and Workloads}

The ALOJA repository provides all the data for learning models of the Hadoop execution and environments. As indicated in sections \ref{sec:background} and \ref{sec:benchmarks} the data-set currently contains up to 40.000 Hadoop executions of 8 different benchmarks from the Intel HiBench suite. Each execution is composed of a prepare Hadoop job that generates the data and a proper benchmark e.g., Teragen and Terasort. Although the job of interest is generally the proper benchmark (i.e. Terasort), prepare jobs are also valid jobs that can be also used for training models. This leaves us with over 80.000 executions to learn from. Each benchmark is run with different configurations, including clusters and VMs, networks, storage drives, internal algorithms and other Hadoop parameters. Table~\ref{table:dataset_properties} summarizes the different properties of the data-set.

\begin{table}[h!tbp]
\centering
\resizebox{0.48\textwidth}{!}{%
\begin{tabular}{|l|l|}
	\hline
	\multicolumn{2}{|c|}{Benchmarks}\\ \hline
	\multicolumn{2}{|c|}{bayes, terasort, sort, wordcount, kmeans, pagerank}\\ 
	\multicolumn{2}{|c|}{dfsioe\_read, dfsioe\_write }\\ \hline \hline
	\multicolumn{2}{|c|}{Hardware Configurations} \\ \hline
	Network & Ethernet, Infiniband \\ \hline
	Storage & SSD, HDD, Remote Disks \{1-3\} \\ \hline
	Cluster & \# Data nodes, VM description \\ \hline \hline
	\multicolumn{2}{|c|}{Software Configurations} \\ \hline
	Maps & 2 to 32 \\ \hline
	I/O Sort Factor & 1 to 100 \\ \hline
	I/O File Buffer & 1KB to 256KB \\ \hline
	Replicas & 1 to 3 \\ \hline
	Block Size & 32MB to 256MB \\ \hline
	Compression Algorithm & None, BZIP2, ZLIB, Snappy \\ \hline
	Hadoop Info & Version \\ \hline
\end{tabular}
}
\vspace{2mm}
\caption{Configuration parameters on the ALOJA data-set}
\vspace{-4mm}
\label{table:dataset_properties}
\end{table}

\begin{figure*}[tbp]
\centering
\footnotesize
\begin{tabular}{|l|l|l|l|l|l|l|l|l|l|}
\hline
id\_exec & id\_cl & bench & exe\_time & start\_time & end\_time & net & disk & bench\_type & maps\\
2 & 3 & terasort & 472.000 & 2014-08-27 13:43:22 & 2014-08-27 13:51:14 & ETH & HDD & HiBench & 8\\
\hline
\multicolumn{10}{c}{~} \\ \hline
iosf & replicas & iofilebuf & compression & blk\_size & \# data nodes & VM\_cores & VM\_ram & validated & version\\
10 & 1 & 65536 & None & 64 & 9 & 10 & 128 & 1 & 1\\
\hline
\end{tabular}
\caption{Example of a logged execution in the ALOJA repository, without its related system profiling data}
\vspace{-3mm}
\label{figure:instance_example}
\end{figure*}

Figure~\ref{figure:instance_example} shows an example of an execution entry in the ALOJA repository. From each entry we distinguish the input variables, those features that are defined before the execution by the user or the environment; the output variables, those features that are result of the execution like the execution time; and other variables providing extra information identifying the execution or added by the ALOJA users like the \emph{validated} field indicating whether a user has reviewed or not that execution. At this time we focus our interest on the elapsed time for a given execution, the execution time, as this can determine the cost of the execution and indicate whether an execution is (not) successful. Another important output variables include resources consumed such as CPU, memory, bandwidth and storage consumption over time. But, as our initial concern is to reduce the number and duration of executions required to characterize (learn) the system behavior, at this time we center our efforts on learning and predicting the execution time for a given benchmark and configuration.

\subsection{The Learning Process}

\begin{figure}[h!tbp]
\centering
\includegraphics[width=0.95\linewidth]{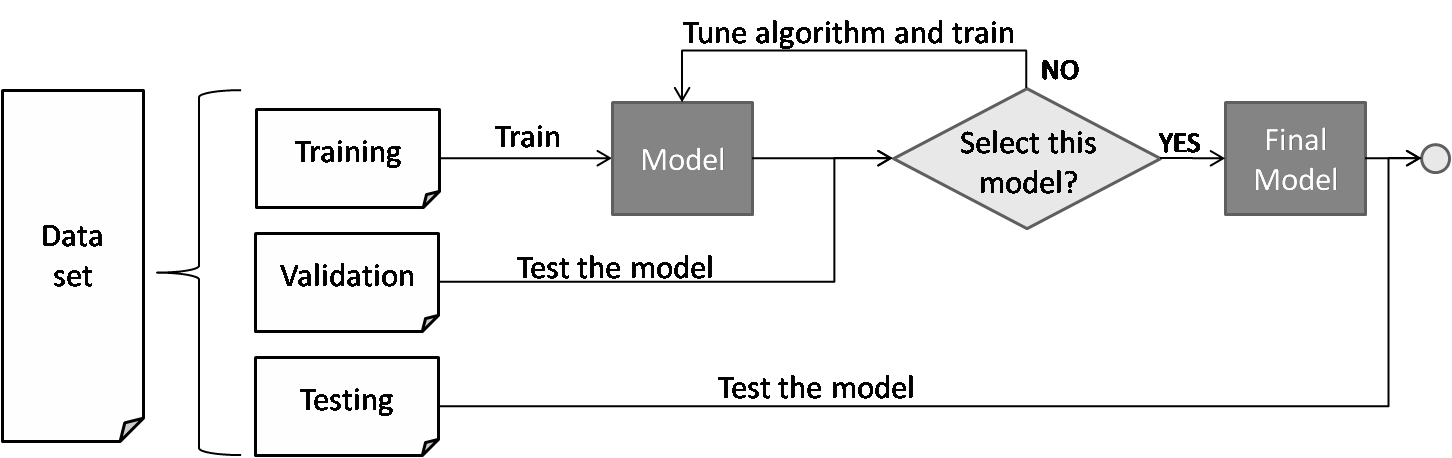}
\caption{Data-set splitting and learning schema}
\label{figure:data-splits-schema}
\end{figure}

The learning methodology is a 3-way step model involving training, validation and testing; see Figure~\ref{figure:data-splits-schema}. The data-set is split (at this time through random sample) and two subsets are used to train a model and validate it. A selected algorithm (taken from those detailed later) learns and characterizes the system, also identifying and retaining the 'best' parameters (testing them on the validation split) from a list of preselected input parameters. The third subset is used to test the best-from-parameters model. All learning algorithms are compared through the same test subset.

At this time, the ALOJA user can choose among four learning methods: Regression Trees, Nearest Neighbors, Feed-Forward Artificial Neural Networks, and Polynomial Regression. Each of these has different mechanisms of learning and has strengths and weaknesses in terms of handling larges volume of data, being resilient to noise, and dealing with complexity. The selected methods are explained as follows:

\textbf{Regression tree} algorithm: we use the M5P~\cite{Quinlan1992,Wang1997} from the RWeka toolkit. The parameter selection (number of instances per branch) is done automatically after comparing iteratively the prediction error of each model on the validation split.

\textbf{Nearest neighbor} algorithm: we use the IBk~\cite{Aha1991}, also from the RWeka toolkit. The number of neighbors is also chosen the same way as parameters on the regression trees.

\textbf{Neural networks}: we use a 1-hidden-layer FFANN from {\em nnet} R package~\cite{nnet:venables2002} with pre-tuned parameters as the complexity of parameter tuning in neural nets require enough error and retrial to not provide a proper usage of the rest of tools of the framework. Improving the usage of neural networks, including the introduction of deep learning techniques, is in the road-map of this project.

\textbf{Polynomial regression}: a baseline method for prediction, from the R core package~\cite{stats:rcore2014}. Experiences with the current data-sets have shown that linear regression and binomial regression do not produce good results, but trinomial approximates well. Higher degrees have been discarded because of the required computation time, also to prevent over-fitting.

\begin{figure}[h!tbp]
\centering
\includegraphics[width=0.95\linewidth]{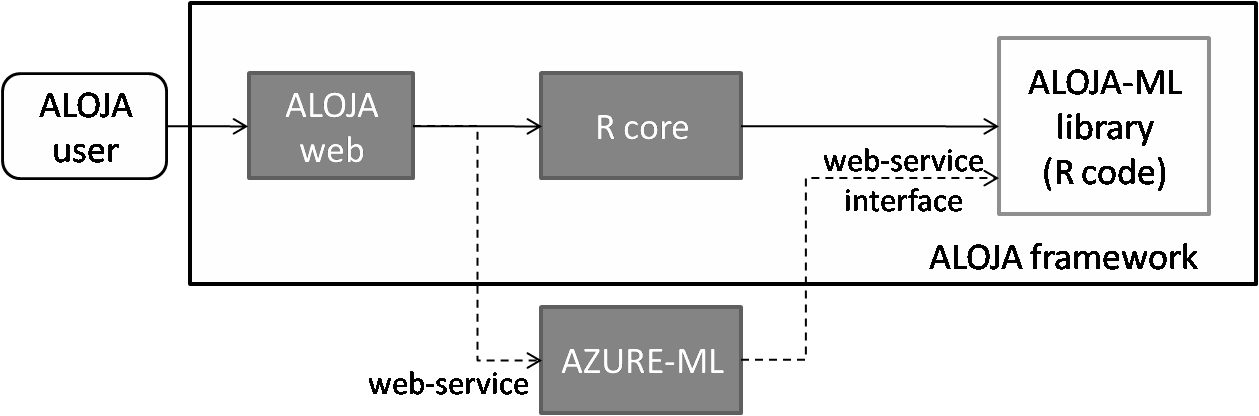}
\caption{Schema of AZURE-ML on the ALOJA framework}
\label{figure:alojaml-azureml}
\end{figure}

The ALOJA-ML tools are implemented in R and are available as a library in our on-line code repository\footnote{https://github.com/Aloja/aloja-ml}. The ALOJA platform loads the library from the web and executes it in the deployed web server, but access can be initiated from any R-based platform. In this way any service or application can call our tool-set for predicting, clustering or treating Hadoop executions. Moreover, our library can be embedded on Microsoft Azure-ML services~\cite{microsoftazureml}, delegating the modeling and prediction process to the cloud, thereby reducing the ALOJA platform code footprint and enabling scaling the Cloud.
A diagram of the ALOJA-ML library is depicted in Figure~\ref{figure:alojaml-azureml}.

\subsection{Modeling Results}

\begin{figure}[h!tbp]
\centering
\includegraphics[width=0.95\linewidth]{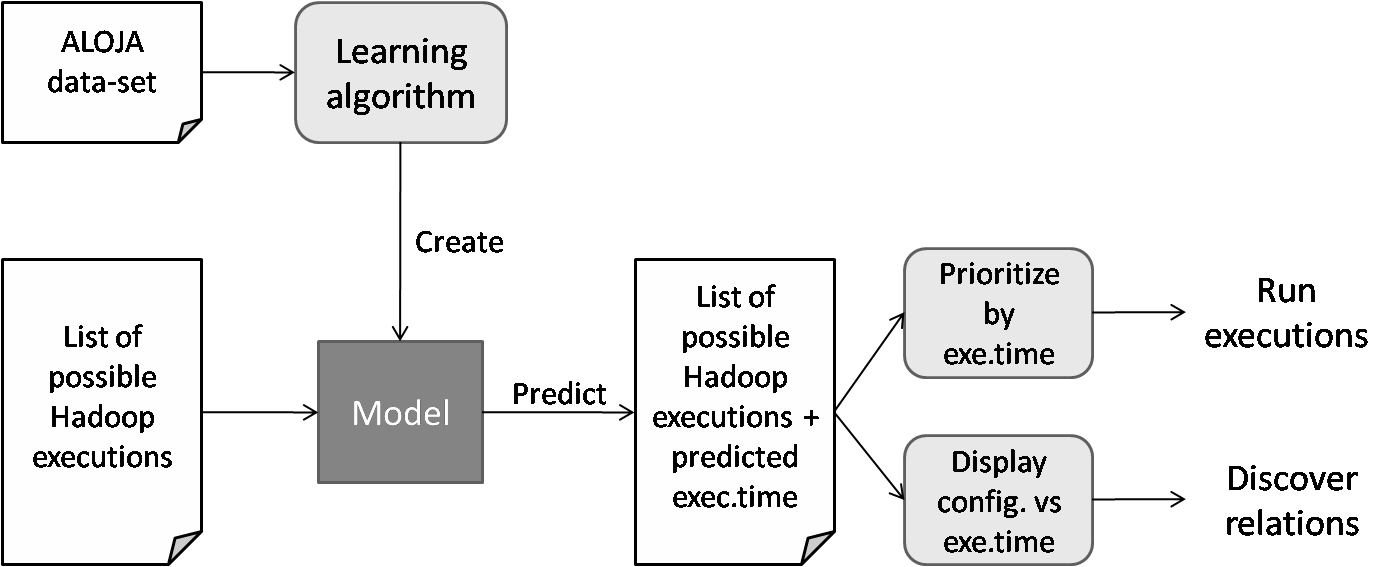}
\caption{Learning and prediction schema}
\label{figure:learn-schema}
\end{figure}

Predicting the execution time for a given benchmark and configuration is the first application of the predictive analytics tool-set. Knowing the expected execution time for a set of possible experiments helps decide which new tasks to launch, their priority order, or just pre-calculate the cost of resource leasing in case that depends on resources per time.
Figure~\ref{figure:learn-schema} shows the learning and prediction data flow.

\subsubsection{Comparing Modeling Algorithms}
As executions cost money and time, we would like to execute as few jobs as possible to model and predict the rest of possible executions. This makes deciding the best sizes for such splits a challenge, as we seek to require as few benchmark executions as possible to train while maintaining good prediction accuracy. This is to build an accurate model from the minimum number of observations. Here we will check the accuracy of predictors given different sizes of training sets, also compare the different selected algorithms, observe how much we can generalize the model to all workloads, and also how ML improves prediction in front of other rule of the thumb techniques applied in the field.

\begin{figure*}[h!tb]
\center
\resizebox{\textwidth}{!}{%
\begin{tabular}{| l | c | c | c | c | c |}
    \hline
    Algorithm & MAE Valid. & RAE Valid. & MAE Test & RAE Test & Best parameters \\ \hline \hline
	Regression Tree & 138.896 & 0.15303 & 401.775 & 0.14885 & min instances per branch = 1 \\ \hline
	Nearest Neighbors & 110.406 & 0.10498 & 248.354 & 0.11209 &  k-nearest neighbors = 3 \\ \hline
	FFA Neural Nets & 135.490 & 0.15271 & 358.195 & 0.16335 & 33:300:1, it1000, $TanH$, $\eta=0.2$, $\mu=0$\\ \hline
	Polynomial Regression & 167.982 & 0.23217 & 354.936 & 0.25414 & degrees = 3\\ \hline
\end{tabular}
}
\caption{Mean and Relative Absolute Error (MAE and RAE) per method, on best split and parameters found}
\vspace{-3mm}
\label{table:learning-error}
\end{figure*}

\begin{figure}[h!tb]
\center
\begin{tabular}{| l | c | c | c |}
    \hline
    Algorithm & 50/25/25 & 37.5/37.5/25 & 20/55/25\\ \hline \hline
	Reg. Trees & 0.15669 & 0.16760 & 0.18225 \\ \hline
	N-Neighbors & 0.11054 & 0.12507 & 0.16433 \\ \hline
	FFA Neural Nets & 0.16335 & 0.15249 & 0.18216 \\ \hline
	Polynomial Reg. & 0.25414 & 0.26025 & 0.90056 \\ \hline
\end{tabular}
\caption{Relative Absolute Error (RAE) per method on test data-set, with different $\%$ splits for Training/Validation/Testing}
\label{table:learning-error-splits}
\end{figure}

As previously said, we have several prediction algorithms from which to create our models, as well as different parameters and choices on the training type. Figure~\ref{table:learning-error} shows the average errors, absolute and relative, for the validation and the testing process for each learning algorithm, using a generous training set (50\% trainnig, 25\% validation and 25\% testing). An interesting thing is to check how much we can reduce the training set without losing too much accuracy, as far as this is possible. Figure~\ref{table:learning-error-splits} shows the training versus validation/test model accuracy, using the best parameters found previously.

As seen there, using regression trees and nearest neighbor techniques we can model and predict the execution time for our Hadoop traces, spending less than a minute to train them. We consider that, with a more dedicated tuning, neural and deep believe networks could improve results, despite requiring around 20 minutes to train with the given data in a commodity computer (single processor Intel i7). After testing, linear and polynomial regressions were set aside as they achieve poor results when compared with the other algorithms, and the time required to generate the model is impractical for the amount of data being analyzed (around an hour of training).

\subsubsection{Generalization of Models}

Another key concern was whether a learned model could be generalized, using data from all the observed benchmarks, or would each execution/benchmark type require its own specific model. One motivation to create a single general model was to reduce the overall number of executions and to generate powerful understanding of all workloads. Also, there was an expectation that our selected algorithms would be capable of distinguishing the main differences among them (e.g., a regression tree can branch different trees for differently behaving benchmarks). On the other hand, we knew that different benchmarks can behave very differently and generalizing might compromise model accuracy. 

Figure~\ref{table:learning-general-specific-2} shows the RAE for passing each benchmark individually through both a general model and a model created using only its type of observations. In almost all cases the specific model fits similarly or better than the general model, and the different algorithms show same trends, but neural networks are capable to fit specific benchmarks much better than the other methods. This can be caused because a general neural net requires more time or examples to learn, but with precise benchmarks is able to perform better.

\begin{figure}[h!tbp]
\center
\begin{tabular}{| l | c | c |}
    \hline
    Benchmark & General Model & Specific Model \\ \hline \hline
	bayes & 0.12624 & 0.05134 \\ \hline
	dfsio\_read & 0.21667 & 0.29965 \\ \hline
	dfsio\_write & 0.19057 & 0.10763 \\ \hline
	k-means & 0.12870 & 0.12842 \\ \hline
	pagerank & 0.10964 & 0.11948 \\ \hline
	sort & 0.20044 & 0.12823 \\ \hline
	terasort & 0.12888 & 0.12599 \\ \hline
	wordcount & 0.20579 & 0.09702 \\ \hline
\end{tabular}
\caption{Comparative RAE for each benchmark, predicting them using the general vs. a fitted model with regression trees}
\label{table:learning-general-specific-2}
\end{figure}

We are concerned about avoiding over-fitting models, as we would like to use them for predicting unseen benchmarks similar to the ones already known in the future. Also, the fact that there are some benchmarks with more executions conditions the general model. After seeing the general vs specific results, we are inclined to use benchmark-specific models in the future, but not discarding using a general one when possible.

After the presented set of experiments and derived ones, we conclude that we can use ML predictors for not only predict execution times of unseen executions, but also for complementing other techniques of our interest, as we present in the following section. Those models provide us more accuracy that techniques used as rules of thumb like Least Squares or Linear Regressions (achieving LS with each attribute an average RAE of 1.90216 and Linear Regression a RAE of 0.70212).

\subsubsection{Applications}

The ALOJA framework incorporates these prediction capabilities in several tools. One of them is predicting the performance of known benchmarks on a new computing cluster, as far as we are able to describe this new cluster, so the new executions for such cluster are automatically decided. Also, having the hardware configuration of such cluster allows us to find the best software configurations for our benchmark executions. In case of a new benchmark entering the system, we can attempt to check if any of the existing models for specific benchmarks fits the new one, and then treat the new benchmark as the known one, or expand or train a new model for this benchmark.

Other important usage, incorporated into the ALOJA framework, is unfolding the space of possible configurations and fill it with predicted values for those configurations without an observed execution. Then, observe the expected importance of a given parameter or a given range of values on a parameter, with reference to performance. Knowing the expected execution time of a Hadoop workload in a given cluster allows the system to schedule jobs better, improving consolidation and de-consolidation processes~\cite{DBLP:conf/icpp/BerralGT13}, and reduce resource consumptions by maintaining any quality of service or job deadline preservation. Or even schedule while tuning parameters for each workload to meet deadlines while preventing resource competition.

\section{Use Cases}
\label{sec:casesofuse}

In this section we present some specific uses for the predicting model, included in the ALOJA platform and with high priority in the road-map of the ALOJA project. These are 1) anomaly detection, by detecting faulty executions through comparing their execution times against the predicted ones; 2) identification of which executions would model best a given Hadoop scenario or benchmark, by clustering the execution observations, and taking the resulting cluster centers as recommended configurations; and 3) identification of which features, for each benchmark or hardware infrastructure, are the most relevant towards speeding up the execution.

\subsection{Anomaly Detection Mechanisms}

An application of our predictive analytics tool-set is to detect anomalous executions or executions to be revised by an operator, and flag them automatically. Detecting automatically executions susceptible of being failures, or even executions not modeled properly, can save time to users who must check each execution, or can require less human intervention on setting up {\em rules of thumb} to decide which executions to discard. 

\begin{figure}[h!tbp]
\centering
\includegraphics[width=0.95\linewidth]{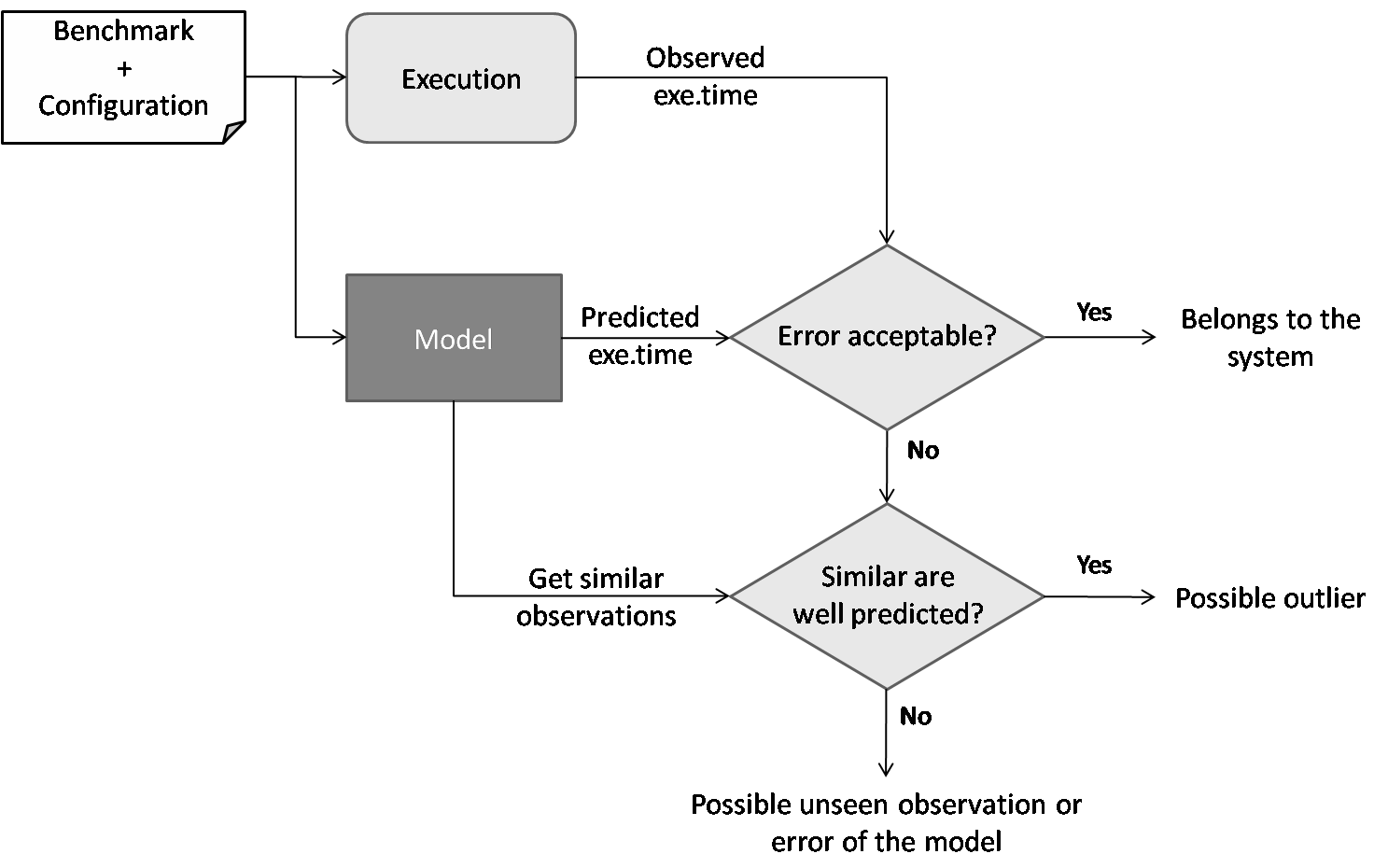}
\vspace{-2mm}
\caption{Anomaly detection schema}
\vspace{-4mm}
\label{figure:outlier-schema}
\end{figure}

Having benchmark behaviors modeled, we can apply model-based anomaly detection methods. After validating a learned model, this can be considered as the `rule that explains the system', and any observation that does not fit into the model (this is, the difference between the observed value and the predicted value is bigger than expected), is considered anomalous. Here we flag anomalous data-set entries as {\em warnings} and {\em outliers}: 1) a {\em warning} is an observation whose error respect to the model is $n$ standard deviations from the average error; 2) an {\em outlier} is a mispredicted observation where other similar observations are well predicted, i.e. a warning that more than a half of its neighbor observations (those ones that differ in less than $h$ attributes, or with Hamming distance $<h$) are well predicted by the model. Figure~\ref{figure:outlier-schema} shows the anomaly decision making schema.

\subsubsection{Validation and Comparisons}

First of all we can auto-analyze the ALOJA data-set, applying the anomaly detection method with the model created from the same data-set, knowing that it can contain anomalous executions. Here we perform two types of experiments, one testing the data of a single benchmark (i.e. Terasort, with 7844 executions) with a model learned from all the observations (i.e. M5P regression tree algorithm), and one testing it against a model created from its specific type of observations.

Applying the method using a model learned from all the observed executions, and adjusting the parameters around $h=\{0...3\}$ and $n=3$, we detected 20 executions from $\sim7800$, all of them having times not matching with what the model would expect given their configuration. After reviewing them manually, we detected that those execution were valid executions, meaning that they finished correctly, but something altered their execution, as other repetitions finished correctly and in time. Further, when learning from Terasort observations only, the more fitted model is able to detect 4 more executions as anomalous, which in the general model where accepted because of similarities with other benchmark executions with similar times. From here on we recommend to validate outlier executions from models trained exclusively with similar executions.

The Hamming distance value, used to detect comparable neighbors, depends on the dispersion of our training examples, if the example executions are very different we should increase the $h$ parameter, allowing comparisons with more examples but reducing the accuracy of the decision. Also for benchmarks or environments with high precise results, the sigma value $n$ can be reduced to detect outliers, but when results are disperse we should consider acceptable errors those between $n \sim 3$.

\begin{figure}[h!tbp]
\centering
\vspace{-3mm}
\includegraphics[width=0.95\linewidth]{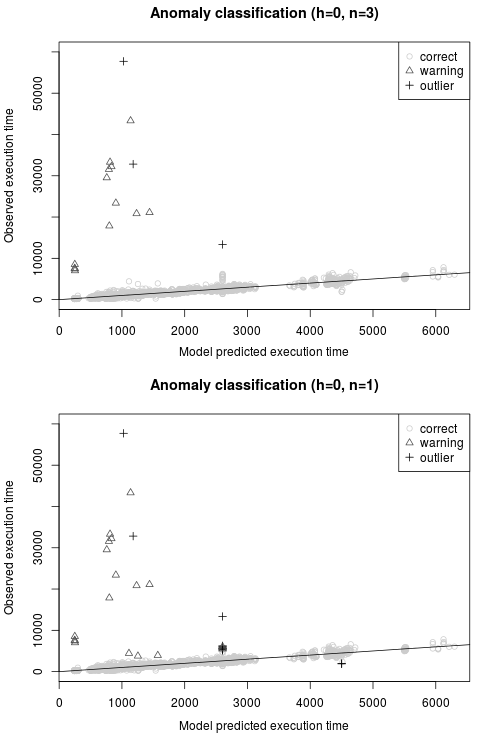}
\vspace{-3mm}
\caption{Automatic outlier detection ($h=0$, $n=\{1,3\}$)}
\label{figure:outlier-results}
\end{figure}

In our tests different values of Hamming distance show low variation, as outliers are easily spotted by neighbors at distance $h=1$ or even identical executions ($h=0$), where the error is usually high. Setting up a distance $h=0$ and $n=3$, where an observation only is an outlier when there are at least two identical instances with an acceptable time, from 24 executions are detected as anomalous, 17 are considered warnings while 7 are considered outliers. Such warnings can be set to revision by a human referee to decide whether are outliers or not. Also, we may decide to reduce the number of directly accepted observations by lowering $n$ (standard deviations from the mean) from 3 to 1. In this situation, we increase slightly the number of detections to 38 (22 warnings and 16 outliers). Figure~\ref{figure:outlier-results} shows the comparative of observed versus predicted execution times, marking outliers and warnings. Also figure~\ref{figure:outlier-confmatrix} shows the confusion matrices for the automatic versus a manual outlier tagging, automatic versus {\em rule of thumb} (semi-automatic method where ALOJA checks if the benchmark has processed all its data), and automatic warnings versus manual classification as incorrect or ``to check'' (which is if the operator suspects that the execution has lasted more than 2x the average of its similar executions).

\begin{figure}[h!tb]
\center
\subfloat{%
\scriptsize
\begin{tabular}{ r | c | c | }
  automatic $\rightarrow$ & & \\ manual $\downarrow$ & Outlier & OK \\ \hline
  Anomaly & 12 & 22 \\ \hline
  Legitimate & 4 & 7786 \\ \hline
\end{tabular}
}%
~
\subfloat{%
\scriptsize
\begin{tabular}{ r | c | c | }
  automatic $\rightarrow$ & & \\ {\em semi-auto.} $\downarrow$ & Outlier & OK \\ \hline
  Anomaly & 7 & 0 \\ \hline
  Legitimate & 9 & 7786 \\ \hline
\end{tabular}
}%
\\
\subfloat{%
\scriptsize
\begin{tabular}{ r | c | c | }
  automatic $\rightarrow$ & & \\ manual $\downarrow$ & Warning & OK \\ \hline
  {\em to check} & 22 & 0 \\ \hline
  Legitimate & 0 & 7786 \\ \hline
\end{tabular}
}%
\caption{Confusion matrices for different methods}
\label{figure:outlier-confmatrix}
\end{figure}

Those confusion matrices show anomalies considered legitimate by the automatic method. After human analysis we discovered that such executions are failed executions with a very low execution time, whose prediction error is also low and the method does not detect them as outlier. Discovering this let us to determine a new manual rule for executions not exceeding a certain amount of time (i.e. a minute), to be marked as possible failed executions.

Finally, this method can be used not only for validating new executions but also to clean our ALOJA data-set and retrain our models without outliers, letting us to discard outlier executions that make our algorithms have low accuracy. In the first regression tree case, shown in the previous subsection, by subtracting the observations marked as outlier we are able to go from a prediction error of $0.15303$ to $0.13561$ on validation, and $0.14885$ to $0.14024$ with the test data-set.

\subsubsection{Use cases}

Spotting failed executions in an automatic way saves time to users, but also let the administrators know when elements of the system are wrong, faulty, or have unexpectedly changed. Further, sets of failed executions with common configuration parameters indicate that it is not a proper configuration for such benchmark; or failing when using specific hardware shows that such hardware should be avoided for those executions.

Also, highlighting anomalous executions make easier to analyze data, even more when having such amount of executions in the repository, plus the +600.000 other performance traces associated to the repository executions. Also it allows to use other learning models less resilient to noise.

\subsection{Recommending Executions}

When modeling a benchmark, a set of configurations or a new hardware set-up, some executions must be performed to observe its new behavior. But as these executions cost money and time, we want to run as few of them as possible. This means running the minimum set of executions that define the system with enough accuracy.

From a model of our executions we can attempt to obtain which of those executions are the most suitable to run on a new system and use the results to model it; or we can use the model to obtain which executions, seen or unseen in our data-set, can be run and used to model. The ALOJA data-set, obtained from random or serial executions, can contain similar executions, introduce redundancy or noise. And find which minimal set of executions are the ones that minimize the amount of training data is a combinatorial problem on a big data-set.

\begin{figure}[h!tbp]
\centering
\includegraphics[width=1\linewidth]{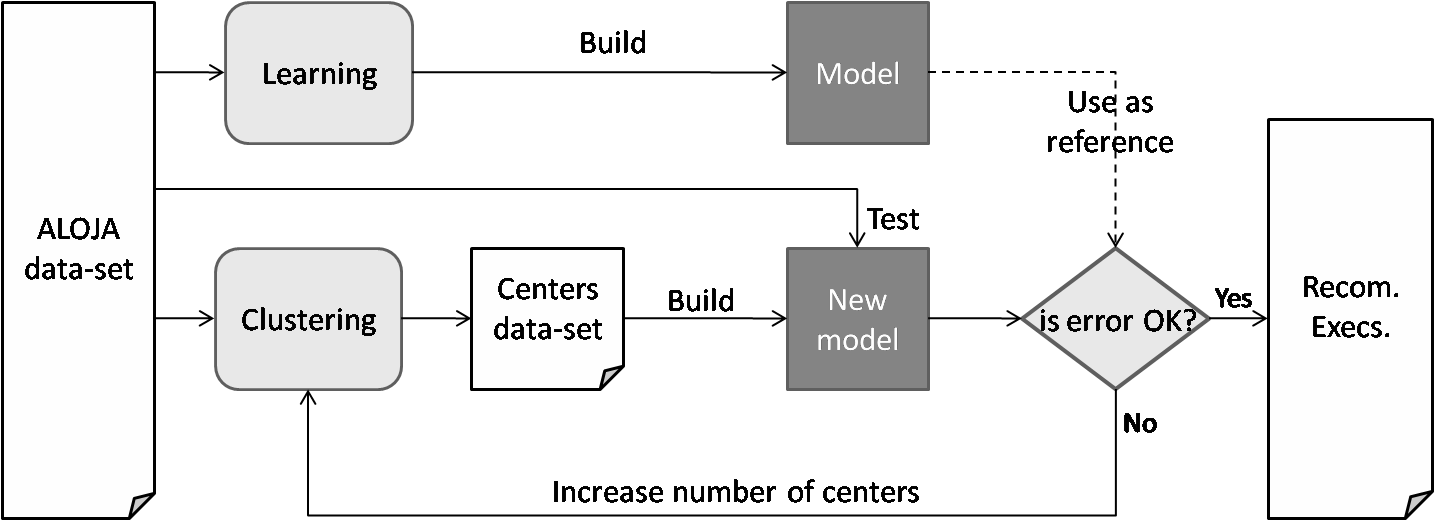}
\caption{Finding recommended executions schema}
\vspace{-4mm}
\label{figure:minconfs-schema}
\end{figure}

One of the methods of the ALOJA-ML tool-set to achieve this, is to cluster our observed executions (i.e., apply the {\em k-means} algorithm~\cite{stats:rcore2014}), obtain for each cluster a representative observation (i.e., its centroid), and use the representatives as the recommended set of executions. Determine the number of centroids (recommendations) required to cover most of the information is the main challenge here. At this time we iterate through a range of $k$, as figure~\ref{figure:minconfs-schema} displays, reconstructing the model with those recommended observations, and testing it against our data-set or against a reference model. From here on, we decide when the error is low enough or when we exceed the number of desired executions.

\subsubsection{Validation and Comparisons}

\begin{figure}[h!tbp]
\centering
\includegraphics[width=0.95\linewidth]{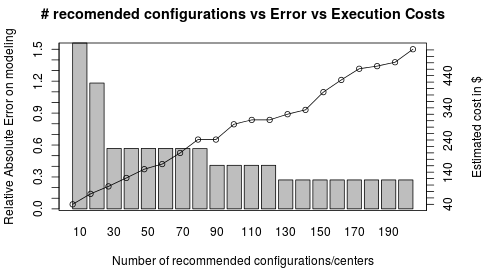}
\caption{Number of recommended executions vs. error in modeling vs. execution cost}
\label{figure:minconfs-k-error-cost}
\end{figure}

For each iteration, looking for $k$ clusters, we compute the error of the resulting model against the reference data-set or model, also we can estimate the cost of running those executions in the clusters we tested, with an average execution cost of $6.85~\$/hour$. Notice that the estimated execution time is from the seen data-set, and applying those configurations on new clusters or unseen components may increase or decrease the values of such estimations, and it should be treated as a guide more than a strict value. Figure~\ref{figure:minconfs-k-error-cost} shows the evolution of the error and the execution cost per each $k$ group of recommendations from our ALOJA data-set. See that more executions implies more accuracy on modeling and predicting, but more cost and execution time. 

Further, to test the method against a new cluster addition, we prepared new a setup (on premise, 8 data nodes, 12 core, 64 RAM, 1 disk), and run some of the recommendations obtained from our current ALOJA data-set. We get 6 groups of recommendations with from $k=\{10...60, step=10\}$, and we executed them in order (first the group of $k=10$ and so on, removing in this case the repeated just to save experimentation time). We found that with only those 150 recommendations we are able to learn a model with good enough accuracy (tested with all the observed executions of the new cluster), compared to the number of executions needed from the ALOJA data-set to learn with similar accuracy.

\begin{figure}[h!tbp]
\centering
\vspace{-3mm}
\includegraphics[width=0.95\linewidth]{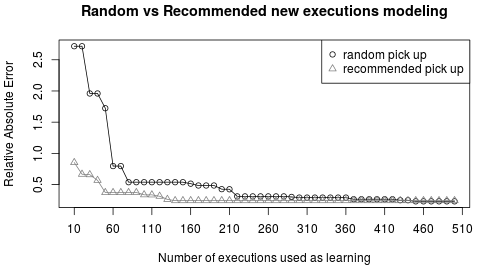}
\vspace{-4mm}
\caption{Random vs. recommended executions}
\vspace{-3mm}
\label{figure:minconfs-learning-rate}
\end{figure}

Figure~\ref{figure:minconfs-learning-rate} shows the comparative of learning a model with $n$ random observations picked from the ALOJA data-set, seeing how introducing new instances to the selected set improves the model accuracy, against picking the new executed instances (this case in order of recommendation groups), and see how it improves the learning rate of the new cluster data-set. We can achieve low prediction errors very quickly, in front of a random execution selection.

\subsubsection{Use cases}

Often executions on a computing cluster are not for free, or the amount of possible configuration (HW and SW) to test are huge. Finding the minimal set of executions to be able to define the behavior of our Hadoop environment helps to save time and/or money. Also, the operator would like to prioritize executions, running first those that provide more information about the system, and then run the rest in descending order of relevance. This is useful when testing or comparing our environment after modifications, sanity checks or validating clone deployments.

Further, when adding new benchmarks or resources it is usual that the new benchmark is similar in behavior to another previously seen, or that a hardware component is similar in behavior to another. Instead of testing it from random executions, we could use the main executions for the most similar seen environments to test it, and although results can not fit well with previous models (in fact the new environment can be different), use the new observed results as a starting point to create a new model. The study of how it performs well against other example selection methods for Hadoop platforms and brand new benchmarks is in the ALOJA road-map for near future research.

\subsection{Ranking Features}

Knowing how variables affect the execution is important, and it can be retrieved from examining the learned model. But ranking the combination of variables from fast to slow, and detecting which variables produce such gradient can be way practical in case the model is not easy to interpret (like neural networks or nearest neighbors). Selecting a subspace of configurations (due to practical reasons), we can predict all configurations tested and not tested, and rank the configurations from slower to faster. Then, we find which variables produce the greatest changes in such ranking, indicating which ones create the greatest divide, and so on for each variable choice.

We can use several methods for ranking variables, like the Gini factor for each variable versus the execution time. Also we can use another method, separating the ranked configurations in a dichotomous way (Algorithm~\ref{alg:dichotomous}), after determining which variable separates better the slow configurations from the faster ones, the algorithm fixes this variable and repeats for each of its distinct values. We found that both methods are useful with different results, so our tool-set implements both for the user to choose and experiment.

Depicted in Figure~\ref{fig:predictions} as an example, we select an on-premise cluster formed by 3 data-nodes with 12 cores/VM and 128GB RAM, and we want to observe the relevance of variables \emph{disk} (local SSD and HDD), \emph{network} (IB and Eth), \emph{IO file buffer} (64KB and 128KB) and \emph{block size} (128, 256) for \emph{terasort}, fixing the other variables (\emph{maps} = 4, \emph{sort factor} = 10, no compression and 1 replica). We train a model for this benchmark, predict all the configurations available for the given scenario, then using the dichotomous algorithm we get the tree of relevant variables.

\begin{algorithm}[h!tb]
\begin{algorithmic}[1]
	\Function{Least.Splits}{$e$}
		\If {$|e| > 1$}
			\State $bv \gets null$ ; $lc \gets \infty$
			\For{$i \in variables(e)$}
				\State $c\gets 0$
				\For{$j \in [2,|e|]$}
					\If{$e[i,j] \neq e[i,j-1]$} \State $c\gets c + 1$ \EndIf
				\EndFor
				\If{$c < lc$}
						\State $\langle bv,lc \rangle \gets \langle i,c \rangle$
				\EndIf
			\EndFor	
			\State $t \gets empty\_tree()$
			\For{$v \in values\_of(e[bv])$}
				\State $branch(t,"bv=v") \gets$ Least.Splits($e[bv = v]$)
			\EndFor
			\State \textbf{return} $t$
		\Else
			\State \textbf{return} $prediction(e)$
		\EndIf
	\EndFunction
\end{algorithmic}
\caption{Least Splits Algorithm}
\label{alg:dichotomous}
\end{algorithm}

\begin{figure}[h!tb]
\centering
\footnotesize
\subfloat{%
    \begin{tabular}{| c | c | c | c | r |}
    \hline
	Network &Storage &IO.File Buffer &Predicted Execution Time (s)\\ \hline \hline
	ETH & HDD & 65536 &	 963.471 \\
	IB & HDD & 65536 & 726.372 \\
	ETH & SSD & 65536 & 330.583 \\
 	IB & SSD & 65536 & 237.817 \\
	ETH & HDD & 131072 & 963.471 \\
	IB & HDD & 131072 & 726.372 \\
	ETH & SSD & 131072 & 330.583 \\
	IB & SSD & 131072 & 237.817 \\ \hline	
	\multicolumn{4}{|c|}{Terasort, 4 maps, sort factor 10, block size 64MB, no comp,}\\
	\multicolumn{4}{|c|}{deployed on a local cluster with HDD/SSD, ETH/IB}\\
    \hline
    \end{tabular}
}%
\hfill
\vspace{1mm}
    \centering
    {
    \begin{verbatim}
    Disk=HDD
        Net=ETH
            IO.FBuf=131072|65536 -> 963 seconds
        Net=IB
            IO.FBuf=131072|65536 -> 726 seconds
    Disk=SSD
        Net=ETH
            IO.FBuf=131072|65536 -> 331 seconds
        Net=IB
            IO.FBuf=131072|65536 -> 238 seconds
    \end{verbatim}
}%
    \caption{Example of estimation of the selected space of search, with the corresponding descriptive tree using the dichotomous method}
    \label{fig:predictions}
\end{figure}


\section{Conclusion}
\label{sec:conclusions}

In this article we described ALOJA-ML, a tool-set for automated modeling and prediction tasks over benchmarking data repositories, part of the ALOJA project. ALOJA-ML identifies key performance properties of the workloads through machine learning, in this case from the Hadoop ecosystem, to predict performance properties for a workload execution on a given set of deployment parameters that have not been explored before in the testing infrastructure. The work presented includes some selected use cases of the ALOJA-ML tool-set in the scope of the ALOJA platform. One of the presented techniques is used to guide and select the most representative runs of an application that needs to be characterized in a deployment, to reduce the number of samples needed. Another technique is to identify anomalies on large sets of job executions, to filter automatically failed runs. The last technique presented is ranking configuration parameters, to decide a new deployment based on the performance versus the available resources.

Through our experiments, we exposed and demonstrated that using our techniques we are able to model and predict Hadoop execution times for given configurations, with a small relative error around 0.20 depending on the executed benchmark. Further we passed out data-set through an automated anomaly detection method, based on our obtained models, with high accuracy respect a manual revision. Also we deployed a new Hadoop cluster, running the recommended executions from our method, and tested the capability of characterizing it with little executions; finding that we can model the new deployment with fewer executions than by randomly selecting test configurations.

The current road-map of ALOJA-ML includes to add new features, improving the ALOJA framework for Hadoop data analysis and knowledge discovery. Among our main interests are learn how to deal with big amounts of data from Hadoop executions to improve the comprehension and management of such platforms. Next steps include study techniques to characterize computation clusters and benchmarks; introduce new input and output variables looking for new sources of information from the system; study in detail compatibilities and relations among configuration and hardware attributes; improve the methods to select features, examples and learning parameters; and add new executions to the ALOJA data-sets, with new deployments and benchmarks, like e.g. BigBench and BigDataBench.

\ifCLASSOPTIONcompsoc
  \section*{Acknowledgments}
\else
  \section*{Acknowledgment}
\fi

{\small This project has received funding from the European Research Council (ERC) under the European Union's Horizon 2020 research and innovation programme (grant agreement No 639595). This work is partially supported by the Ministry of Economy of Spain under contracts TIN2012-34557 and 2014SGR1051.}

\ifCLASSOPTIONcaptionsoff
  \newpage
\fi

%

\vspace{-1mm}
\begin{IEEEbiography}[{\includegraphics[width=1in,height=1.25in,clip,keepaspectratio]{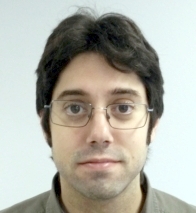}}]{Josep Ll. Berral} received his degree in Informatics (2007), M.Sc in Computer Architecture (2008), and Ph.D. at BarcelonaTech-UPC, speciality on Computer Science (2013). He is a data scientist, working in applications of data mining and machine learning on data-center and cloud environments at UPC, also associate researcher at the Barcelona Supercomputing Center (BSC). He has worked at the High Performance Computing group at the Computer Architecture Department-UPC, also at the Relational Algorithms, Complexity and Learning group at the Computer Science Department-UPC.
\end{IEEEbiography}

\begin{IEEEbiography}[{\includegraphics[width=1in,height=1.25in,clip,keepaspectratio]{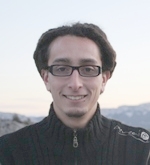}}]{Nicolas Poggi} is a senior researcher with focus on performance and scalability of Data intensive applications. He is currently leading a research project on upcoming architectures for intensive data applications (BigData) at the Barcelona Supercomputing Center (BSC) and Microsoft Research joint center. Nicolas received his PhD in 2014 at the Computer Architecture Department at BarcelonaTech-UPC, where he also obtained his M.Sc in 2007. He received his IT Engineering degree (best student award) with a minor in Business Administration at the American University (UA) in 2005. He is also part of the High Performance Computing group at DAC and the Autonomic Systems and e-Business Platforms research group at BSC.
\end{IEEEbiography}
\vspace{-5mm}
\begin{IEEEbiography}[{\includegraphics[width=1in,height=1.25in,clip,keepaspectratio]{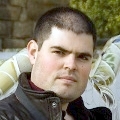}}]{David Carrera} received the MS degree at BarcelonaTech-UPC in 2002 and his PhD from the same university in 2008. He is an associate professor at the Computer Architecture Department of the UPC. He is also an associate researcher in the Barcelona Supercomputing Center (BSC) within the “Autonomic Systems and eBusiness Platforms” research line. His research interests are focused on the performance management of data center workloads. He has been involved in several EU and industrial research projects. In 2015 he was awarded an ERC Starting Grant for the project HiEST. He received an IBM Faculty Award in 2010. He is an IEEE member.
\end{IEEEbiography}
\vspace{-5mm}
\begin{IEEEbiography}[{\includegraphics[width=1in,height=1.25in,clip,keepaspectratio]{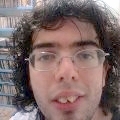}}]{Aaron Call} received his BS at Universitat Politècnica de Catalunya (UPC) in 2014. He is a research support engineer at BSC within the “Autonomic Systems and eBusiness Platforms" research group. He is currently working on the Aloja project of BSC-Microsoft Research. His research interests are Big Data and distributed computing, with special focus on grid computing and networks.
\end{IEEEbiography}
\vspace{-5mm}
\begin{IEEEbiography}[{\includegraphics[width=1in,height=1.25in,clip,keepaspectratio]{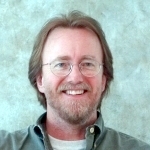}}]{Rob Reinauer} is a Systems Architect in the Microsoft SQL Server Data Warehouse Appliance group where he leads Microsoft contributions to the BSC-Microsoft Research Center and leads efforts in designing high performance MPP Data Warehouse systems for both Appliances on premises and cloud based deployments. Previous to these roles Rob managed the Microsoft SQL Server Systems group, the Microsoft SQL Server Engine organization and held senior technical \& management roles at Pervasive Software, IBM and Tandem computers. Rob Reinauer is the named inventor on multiple US and International patents as well as the author and co-author of numerous publications on computer and networking architecture, performance and design.
\end{IEEEbiography}
\vspace{-5mm}
\begin{IEEEbiography}[{\includegraphics[width=1in,height=1.25in,clip,keepaspectratio]{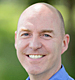}}]{Daron Green} is the Senior Director of Microsoft External Research responsible for external engagement and investment strategy. His BSc is in Chemical Physics (1989, Sheffield) and his PhD in molecular simulation of fluid mixtures (1992, Sheffield). He moved into HPC, being responsible for some European Commission programs and major procurements for the UK Research Councils and Defense clients. He worked for SGI/Cray helping to set up the European Professional Services organization, also worked for Selectica Inc. IBM invited him to help establish its early Grid strategy and he moved to the United States with IBM to form IBM’s Grid EBO. Then he joined Microsoft Research from BT where he was responsible for all sector-based propositions in BT’s Global Services.
\end{IEEEbiography}






\vfill


\end{document}